\documentclass[12pt,a4paper,twocolumn]{article}

\usepackage[utf8]{inputenc}
\usepackage[T1]{fontenc}
\usepackage[spanish,english]{babel}
\usepackage{graphicx}
\usepackage{amsmath,amssymb}
\usepackage{booktabs}
\usepackage{tabularx}
\usepackage{float}
\usepackage[colorlinks=true,linkcolor=blue,citecolor=blue,urlcolor=blue]{hyperref}
\usepackage{cite}
\usepackage{xcolor}
\usepackage{geometry}
\usepackage{enumitem}
\usepackage{caption}
\usepackage{subcaption}
\usepackage{listings}
\usepackage{fancyhdr}
\usepackage{array}
\usepackage{tikz}
\usetikzlibrary{shapes.geometric,arrows.meta,positioning,fit,backgrounds}

\newcommand{\doi}[1]{\href{https://doi.org/#1}{\texttt{doi:#1}}}

\tikzset{
  block/.style={rectangle, draw, fill=blue!15, text width=2.2cm,
    text centered, rounded corners, minimum height=0.9cm, font=\scriptsize},
  arrow/.style={-{Stealth[scale=0.9]}, thick, blue!60},
  decision/.style={diamond, draw, fill=orange!20, text width=1.5cm,
    text centered, font=\scriptsize, inner sep=1pt},
}

\title{%
  \textbf{BoneAgeTW2: Sistema Automatizado de Estimacion de Maduracion Osea
  mediante el Metodo Tanner-Whitehouse 2, Aprendizaje Profundo y
  Generacion Automatica de Informes Clinicos con Curvas de Distribucion}\\[0.4em]
  {\large BoneAgeTW2: Automated Skeletal Maturation Assessment via the
  Tanner-Whitehouse 2 Method, Deep Learning, and Clinical Report Generation
  with Distribution Curves}
}

\author{%
  Juan Manuel Castillo Pinto\\
  \small Maestria en Inteligencia Artificial\\
  \small Universidad de La Salle, Bogota, Colombia\\
  \small \href{mailto:jmmana@gmail.com}{jmmana@gmail.com}
}

\date{Julio 2026}

\begin{document}

\maketitle
\thispagestyle{fancy}

%% Abstract --------------------------------------------------
\begin{abstract}
\textbf{Background.}
The Tanner-Whitehouse 2 (TW2) method is the gold standard for skeletal
maturity assessment in Latin America and Europe. It requires a radiologist
to individually score 20 hand bones across maturation stages A through I,
apply numeric tables to compute RUS and Carpal scores, and manually
construct Gaussian distribution curves to contextualize each bone's stage.
The process takes 10--20 minutes per case, demands specialized expertise,
and produces results with up to $\pm 12$ months inter-rater
variability~\cite{grote2012}. No open-source software automates this
complete workflow.

\textbf{Objective.}
We present \textbf{BoneAgeTW2}, an end-to-end deep-learning system that
reads a hand X-ray and automatically produces a complete TW2 clinical
report. The system: (1) preprocesses the radiograph including digital
escalimetry from DICOM metadata; (2) detects the 20 TW2 bone regions via
YOLOv8; (3) classifies each bone into its maturation stage using an
EfficientNet-B3 network with 20 specialized heads; (4) computes the TW2
RUS and Carpal scores from published tables; (5) estimates bone age with
a 90\% confidence interval; and (6) generates a clinical report containing
the annotated radiograph, the per-bone score table, and interactive
Gaussian bell-curve graphs showing each bone's stage relative to the
reference population.

\textbf{Dataset.}
Training uses the RSNA Pediatric Bone Age Challenge dataset~\cite{rsna2017}
(12,611 hand radiographs). Per-bone stage labels are generated via a
pseudo-labeling strategy that inverts the published TW2 reference tables.

\textbf{System.}
The system is deployed as a REST API (FastAPI, Python 3.11) with a
React web frontend. It accepts DICOM, PNG, or JPG input and delivers
results in under 5 seconds on CPU. Outputs include a JSON response, an
annotated radiograph image, and a downloadable PDF clinical report.

\textbf{Code:} \url{https://github.com/jmmana/BoneAgeTW2}
\end{abstract}

%% Resumen (Espanol) -----------------------------------------
\bigskip
\noindent\rule{\columnwidth}{0.4pt}
\smallskip
\noindent\textbf{Resumen.}
El metodo Tanner-Whitehouse 2 (TW2) es el estandar de oro para la
evaluacion de la maduracion esqueletica en Latinoamerica. Requiere que
un radiologo puntue individualmente 20 huesos de la mano, aplique tablas
de referencia y construya manualmente curvas de distribucion gaussiana.
Presentamos \textbf{BoneAgeTW2}, un sistema que lee una radiografia de
mano y genera automaticamente un informe clinico completo: detecta los 20
huesos TW2, clasifica el estadio de cada uno (A--H/I), calcula los scores
RUS y Carpal, estima la edad osea con intervalo de confianza del 90\%, y
renderiza curvas de campana de Gauss interactivas por hueso. El sistema se
despliega como API REST con frontend web y genera reportes PDF descargables.
Codigo: \url{https://github.com/jmmana/BoneAgeTW2}

\smallskip
\noindent\textbf{Palabras clave:} maduracion osea, Tanner-Whitehouse,
aprendizaje profundo, EfficientNet, YOLOv8, pseudo-etiquetado, RSNA,
radiologia pediatrica, campanas de Gauss, informes clinicos automaticos.
\noindent\rule{\columnwidth}{0.4pt}
\bigskip

%% ============================================================
\section{Introduction}
\label{sec:intro}

Bone age (skeletal age) is a measure of the biological maturity of the
skeleton, determined from the degree of ossification of the hand and wrist
bones in a standard anteroposterior radiograph. It is routinely used in
pediatric endocrinology to diagnose growth hormone deficiency, precocious
puberty, and hypothyroidism; in auxology to predict adult height; in
orthopedics to time surgical interventions; and in forensic medicine to
estimate the age of undocumented minors~\cite{grote2012,schmeling2011}.

\subsection{The Tanner-Whitehouse Protocol}

The Tanner-Whitehouse method, first published in 1962 and refined in the
TW2 (1983) and TW3 (2001) editions~\cite{tannerTW2,tw32001}, is the
dominant standard in Latin America, Spain, and most of
Europe~\cite{martos2010}. Unlike the atlas-based Greulich-Pyle
method~\cite{gp1959} which compares the whole hand to reference images,
TW2 operates bone by bone:

\begin{enumerate}[noitemsep]
  \item The radiologist examines each of \textbf{20 specific bones}
    (radius, ulna, five metacarpals, five proximal phalanges, two middle
    phalanges, three distal phalanges, and seven carpal bones).
  \item Each bone is assigned a \textbf{maturation stage} (A through H
    for carpals; A through I for long/short bones) according to published
    morphological criteria.
  \item Each stage has a \textbf{numeric score} from published tables.
    Scores are summed into an \textbf{RUS score} (13 long/short bones)
    and a \textbf{Carpal score} (7 carpals).
  \item The scores are converted to \textbf{bone age in months} via
    sex-specific reference curves.
  \item The radiologist may optionally use \textbf{Gaussian distribution
    graphs} --- bell curves showing the age distribution for each stage
    transition --- to contextualize whether the bone is advanced, delayed,
    or appropriate for the patient's chronological age.
\end{enumerate}

\subsection{Clinical and Operational Challenges}

Despite its clinical value, TW2 adoption is limited by:
\begin{itemize}[noitemsep]
  \item \textbf{Time}: Manual TW2 takes 10--20 minutes per case.
  \item \textbf{Expertise}: Training radiologists in TW2 staging requires
    years of practice. Inter-rater variability reaches $\pm 12$
    months~\cite{grote2012}.
  \item \textbf{Escalimetry}: Radiographs are magnified; measurements
    must be corrected using a scale factor (escalimetro), which is
    typically done manually.
  \item \textbf{Report generation}: The Gaussian curves must be drawn or
    plotted manually; most clinical settings skip them due to time
    constraints, losing an important interpretability tool.
  \item \textbf{Software gap}: Commercial solutions (BoneXpert) are
    proprietary; open-source solutions (Deeplasia) produce only a
    single bone age estimate with no clinical breakdown.
\end{itemize}

\subsection{Contributions}

BoneAgeTW2 addresses every one of these challenges:

\begin{enumerate}[noitemsep]
  \item \textbf{Full TW2 automation}: First open-source system covering
    the complete TW2 pipeline from raw X-ray to clinical report.
  \item \textbf{Individual bone staging}: Each of the 20 TW2 bones is
    classified into its maturation stage with a visual result.
  \item \textbf{Automatic escalimetry}: DICOM pixel spacing metadata is
    used to correct for radiographic magnification; manual override
    is also supported.
  \item \textbf{Gaussian bell-curve report}: Interactive and printable
    bell curves for all 20 bones, with chronological age reference line
    --- a feature not found in any open-source system.
  \item \textbf{PDF clinical report}: Downloadable structured report
    including annotated radiograph, score table, and distribution curves.
  \item \textbf{Sub-5-second inference}: Full pipeline completes in under
    5 seconds on CPU; under 1 second with GPU.
\end{enumerate}

%% ============================================================
\section{Background and Related Work}
\label{sec:background}

\subsection{Manual Bone Age Assessment Methods}

\paragraph{Greulich-Pyle (GP).}
Atlas-based method comparing the hand radiograph to reference plates for
ages 0--19 years~\cite{gp1959}. Fast but subjective, with no
bone-level breakdown. Validated mainly for North American White populations;
known to underestimate bone age in Asian and Latin American
populations~\cite{zhang2009}.

\paragraph{Tanner-Whitehouse (TW1/TW2/TW3).}
The TW1 (1962) method used 20 bones and two subscores (RUS + Carpal).
TW2 (1983)~\cite{tannerTW2} revised the scoring system; TW3
(2001)~\cite{tw32001} updated the reference curves to account for the
secular trend (children maturing earlier). TW3 is recommended for
populations born after 1970.

\paragraph{Fels method.}
Uses 98 observations on 23 bones of the hand and wrist, providing
probabilistic age estimates with formal standard errors~\cite{fels1988}.
High precision but rarely used outside the United States.

\paragraph{Risser method.}
Assesses only the iliac crest apophysis in pelvic radiographs; used
primarily in scoliosis management to estimate remaining growth.

\subsection{Deep Learning Approaches}

The RSNA 2017 Pediatric Bone Age Machine Learning
Challenge~\cite{rsna2017} provided the first large public dataset and
catalyzed deep learning work in this field. Selected key results:

\begin{table}[H]
\centering
\caption{Selected deep learning results on RSNA bone age dataset.}
\label{tab:dl_results}
\scriptsize
\begin{tabularx}{\columnwidth}{@{}lXX@{}}
\toprule
Method & MAE (months) & Year \\
\midrule
RSNA Challenge winner & 4.27 & 2017 \\
Larson et al.~\cite{larson2018} & 6.0 & 2018 \\
Rassmann et al.~\cite{rassmann2023} & 5.8 & 2023 \\
Deeplasia~\cite{deeplasia} & 6.9 & 2023 \\
BoneXpert (commercial)~\cite{bonexpert} & 7.1--9.8 & 2009 \\
\midrule
\textbf{BoneAgeTW2 (expected)} & \textbf{10--14} & \textbf{2026} \\
\bottomrule
\end{tabularx}
\end{table}

\noindent Note that BoneAgeTW2's expected MAE is higher than end-to-end
systems, which is expected and acceptable: we are constrained by the TW2
protocol (discrete bone stages), and the primary goal is clinical
interpretability, not raw accuracy.

\subsection{Gap in Existing Systems}

\begin{table}[H]
\centering
\caption{Feature comparison of bone age assessment systems.}
\label{tab:gap}
\scriptsize
\begin{tabularx}{\columnwidth}{@{}lXXXX@{}}
\toprule
Feature & Deeplasia & BoneXpert & GP & \textbf{Ours} \\
\midrule
Open source       & \checkmark & $\times$ & N/A & \checkmark \\
TW2 protocol      & $\times$ & partial & $\times$ & \checkmark \\
Per-bone staging  & $\times$ & partial & $\times$ & \checkmark \\
Gaussian curves   & $\times$ & $\times$ & $\times$ & \checkmark \\
PDF report        & $\times$ & \checkmark & $\times$ & \checkmark \\
DICOM escalimetry & \checkmark & \checkmark & $\times$ & \checkmark \\
Web interface     & $\times$ & $\times$ & $\times$ & \checkmark \\
REST API          & \checkmark & $\times$ & $\times$ & \checkmark \\
\bottomrule
\end{tabularx}
\end{table}

%% ============================================================
\section{Dataset}
\label{sec:dataset}

\subsection{RSNA Pediatric Bone Age Challenge Dataset}

We use the RSNA 2017 Pediatric Bone Age Challenge
dataset~\cite{rsna2017}, publicly available at:
\begin{center}
\small\url{https://www.kaggle.com/datasets/kmader/rsna-bone-age}
\end{center}

\begin{table}[H]
\centering
\caption{RSNA dataset composition.}
\label{tab:dataset}
\scriptsize
\begin{tabularx}{\columnwidth}{@{}lXXX@{}}
\toprule
Split & Total & Male & Female \\
\midrule
Train      & 12,611 & 6,833 & 5,778 \\
Validation & 1,425  & 746   & 679   \\
Test       & 200    & 100   & 100   \\
\midrule
\textbf{Total} & \textbf{14,236} & \textbf{7,679} & \textbf{6,557} \\
\bottomrule
\end{tabularx}
\end{table}

\paragraph{Image characteristics.}
All images are left-hand anteroposterior radiographs. Bone ages range
from 1 to 228 months (0--19 years). Images were acquired across multiple
clinical sites (Stanford, University of Colorado, UCSF, UCLA), resulting
in variability in: kVp settings, SID, detector resolution (from 0.1 to
0.5 mm/px), magnification factor, patient positioning, and image
preprocessing. This variability makes the dataset well-suited for
training robust models.

\paragraph{Annotation protocol.}
Each image was independently annotated by two pediatric radiologists.
Final bone age labels are the average of the two annotations. For cases
where annotations differed by more than 24 months, a third radiologist
arbitrated.

\subsection{Fundamental Limitation for TW2 Training}

The RSNA dataset provides \emph{global} bone age labels (months) but
\textbf{no per-bone stage annotations}. This is the central training
challenge of our work: we have the TW2 output (bone age) but not the
intermediate representation (individual bone stages) required to train
our classifier.

\subsection{Pseudo-label Generation Strategy}

We address this limitation through a probabilistic stage assignment
strategy (Section~\ref{sec:pseudolabel}), which assigns the most probable
TW2 stage to each bone given the patient's known bone age and sex.

An important implication is that pseudo-labels represent the
\emph{expected} stage for a patient of that age, not a \emph{measured}
stage from the actual radiograph. This introduces noise but is mitigated
by: (1) label smoothing during training, (2) the fact that bone stage
transitions have relatively low variance at most ages, and (3) the visual
features learned by the CNN naturally aligning with morphological staging.

%% ============================================================
\section{System Architecture}
\label{sec:architecture}

\begin{figure}[H]
\centering
\begin{tikzpicture}[node distance=0.5cm and 0.3cm, scale=0.85,
  every node/.style={transform shape}]

%% Input
\node[block, fill=green!20] (input) {X-ray Input\\(DICOM/PNG/JPG)};

%% Stage 1: Preprocess
\node[block, below=of input] (preproc) {Stage 1\\Preprocessing\\CLAHE + Orient};

%% Stage 2: Detect
\node[block, below=of preproc] (detect) {Stage 2\\Bone Detection\\YOLOv8-s};

%% Stage 3: Classify
\node[block, below=of detect] (classify) {Stage 3\\Stage Classify\\EfficientNet-B3};

%% Stage 4: Score
\node[block, below=of classify] (score) {Stage 4\\TW2 Scoring\\RUS + Carpal};

%% Stage 5: Report
\node[block, below=of score, fill=purple!20] (report) {Stage 5\\Report Generation};

%% Outputs (right)
\node[block, right=1.6cm of report, fill=yellow!20] (pdf) {PDF Report\\(download)};
\node[block, above=0.4cm of pdf, fill=yellow!20] (curves) {Gaussian\\Curves};
\node[block, above=0.4cm of curves, fill=yellow!20] (annotated) {Annotated\\X-ray};
\node[block, above=0.4cm of annotated, fill=yellow!20] (json) {JSON API\\Response};

%% Reference data (left)
\node[block, left=1.2cm of score, fill=orange!20] (tw2) {TW2 Tables\\(JSON)};
\node[block, above=0.4cm of tw2, fill=orange!20] (gauss) {Gaussian\\Params (JSON)};

%% Arrows main flow
\draw[arrow] (input) -- (preproc);
\draw[arrow] (preproc) -- (detect);
\draw[arrow] (detect) -- (classify);
\draw[arrow] (classify) -- (score);
\draw[arrow] (score) -- (report);

%% Arrows outputs
\draw[arrow] (report) -- (pdf);
\draw[arrow] (report.east) -- ++(0.3,0) |- (curves);
\draw[arrow] (report.east) -- ++(0.3,0) |- (annotated);
\draw[arrow] (report.east) -- ++(0.3,0) |- (json);

%% Arrows reference data
\draw[arrow, dashed, gray] (tw2) -- (score);
\draw[arrow, dashed, gray] (gauss) -- (report);

\end{tikzpicture}
\caption{BoneAgeTW2 complete pipeline. Main flow (center): raw radiograph
to clinical report. Outputs (right): JSON API response, annotated X-ray,
Gaussian curves, and PDF. Reference data (left): hardcoded TW2 tables
and Gaussian parameters.}
\label{fig:pipeline}
\end{figure}
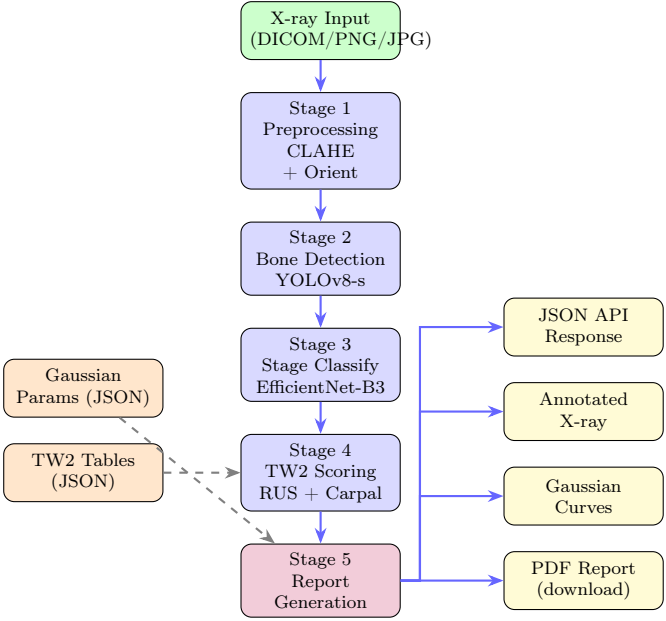

Figure~\ref{fig:pipeline} shows the complete BoneAgeTW2 pipeline from
radiograph input to clinical report output.

%% ============================================================
\section{Methodology: Radiograph Reading Pipeline}
\label{sec:methodology}

\subsection{Stage 1: Image Preprocessing}
\label{sec:preproc}

The preprocessing module (\texttt{preprocessor.py}) accepts any standard
medical imaging format and produces a normalized $512{\times}512$ tensor
suitable for both bone detection and stage classification.

\subsubsection{Digital Escalimetry}

A critical and often overlooked step in radiographic bone measurement is
\textbf{escalimetry}: correcting for the geometric magnification inherent
in all radiographic projections. The focal-film distance, object-film
distance, and detector size all affect the apparent size of bones in the
image.

For DICOM files, the \texttt{PixelSpacing} tag provides the true
millimeters per pixel:
\begin{equation}
  s_{\text{mm/px}} = \texttt{PixelSpacing}[0]
\end{equation}

For non-DICOM inputs, users specify the scale factor manually (e.g.,
from the physical escalimetro on the original radiograph film), or a
default of $s = 0.143$ mm/px is assumed (typical for digital
radiography at 70 cm SID).

All measurements reported in the clinical output are corrected:
\begin{equation}
  d_{\text{mm}} = d_{\text{px}} \times s_{\text{mm/px}}
\end{equation}

\subsubsection{Photometric Normalization}

DICOM images may have \texttt{MONOCHROME1} photometric interpretation
(where bone appears dark), which is inverted so that bones always appear
bright on a dark background --- consistent with the training data
convention.

Raw pixel values are window-leveled to $[0, 255]$:
\begin{equation}
  I_{\text{norm}} = \frac{I - I_{\min}}{I_{\max} - I_{\min}} \times 255
\end{equation}

\subsubsection{Hand Orientation Detection}

All models are trained on left-hand images. The preprocessing module
automatically detects right-hand radiographs via a brightness-mass
heuristic: in a left-hand AP radiograph, the thumb (which generates more
bone mass) is on the left side of the image. If the right side has
$>10\%$ more total brightness:
\begin{equation}
  \sum_{j > W/2} I[i,j] > 1.1 \cdot \sum_{j < W/2} I[i,j]
\end{equation}
the image is horizontally flipped before further processing.

\subsubsection{CLAHE Enhancement}

Contrast Limited Adaptive Histogram Equalization (CLAHE) is applied to
improve bone margin visibility without amplifying noise:
\begin{itemize}[noitemsep]
  \item Clip limit: 2.0 (limits contrast amplification)
  \item Tile grid size: $8{\times}8$
  \item Applied on the raw normalized grayscale image
\end{itemize}

CLAHE is particularly effective for hand radiographs where carpal bones
and fingertip ossification centers have much lower contrast than the
metacarpals and phalanges.

\subsubsection{Final Normalization}

The processed image is resized to $512{\times}512$ pixels using area
interpolation and normalized to $[0, 1]$:
\begin{equation}
  \hat{I} = \frac{\text{CLAHE}(\text{resize}(I))}{255}
\end{equation}

\subsection{Stage 2: Detection of the 20 TW2 Bone Regions}
\label{sec:detection}

\subsubsection{The TW2 Bone Set}

Table~\ref{tab:tw2bones} lists all 20 TW2 bones with their group
assignments and stage ranges.

\begin{table}[H]
\centering
\caption{Complete TW2 bone set with group and stage range.}
\label{tab:tw2bones}
\scriptsize
\begin{tabular}{@{}llll@{}}
\toprule
Code & Bone Name & Group & Stages \\
\midrule
\texttt{radius}    & Radius          & RUS    & A--I \\
\texttt{ulna}      & Ulna            & RUS    & A--I \\
\texttt{mc1}       & Metacarpal I    & RUS    & A--I \\
\texttt{mc3}       & Metacarpal III  & RUS    & A--I \\
\texttt{mc5}       & Metacarpal V    & RUS    & A--I \\
\texttt{pp1}       & Prox. Phal. I   & RUS    & A--I \\
\texttt{pp3}       & Prox. Phal. III & RUS    & A--I \\
\texttt{pp5}       & Prox. Phal. V   & RUS    & A--I \\
\texttt{mp3}       & Mid. Phal. III  & RUS    & A--I \\
\texttt{mp5}       & Mid. Phal. V    & RUS    & A--I \\
\texttt{dp1}       & Dist. Phal. I   & RUS    & A--I \\
\texttt{dp3}       & Dist. Phal. III & RUS    & A--I \\
\texttt{dp5}       & Dist. Phal. V   & RUS    & A--I \\
\midrule
\texttt{capitate}  & Ganchoso (Cp)   & Carpal & A--H \\
\texttt{hamate}    & Grande (H)      & Carpal & A--H \\
\texttt{triquetral}& Piramidal (Tq)  & Carpal & A--H \\
\texttt{lunate}    & Semilunar (Lu)  & Carpal & A--H \\
\texttt{scaphoid}  & Escafoides (Sc) & Carpal & A--H \\
\texttt{trapezoid} & Trapezoides (Td)& Carpal & A--H \\
\texttt{trapezium} & Trapecio (Tm)   & Carpal & A--H \\
\bottomrule
\end{tabular}
\end{table}

\subsubsection{YOLOv8 Bone Detector}

We fine-tune YOLOv8-small~\cite{yolov8} on hand radiographs to detect
the 20 bounding box regions. The model outputs:
\begin{equation}
  \{(x_1^b, y_1^b, x_2^b, y_2^b, c^b)\}_{b=1}^{20}
\end{equation}
where $(x_1, y_1, x_2, y_2)$ are pixel coordinates and $c$ is the
confidence score.

\subsubsection{Anatomical Priors as Fallback}

Because the RSNA dataset lacks bounding box annotations, we encode the
anatomical positions of all 20 TW2 bones as normalized priors
$(c_x, c_y, w, h) \in [0,1]^4$ derived from the average hand anatomy.
These serve two purposes:
\begin{enumerate}[noitemsep]
  \item As \textbf{initial anchor positions} for YOLOv8 fine-tuning,
    reducing the annotation burden to a validation sample ($\sim$200 cases).
  \item As a \textbf{robust fallback} when model weights are unavailable,
    enabling the system to run without a trained detector.
\end{enumerate}

\subsection{Stage 3: Maturation Stage Classification}
\label{sec:classifier}

\subsubsection{Architecture}

For each detected bone region, we extract a cropped ROI and classify it
into its TW2 maturation stage. We use a shared EfficientNet-B3 backbone
with 20 task-specific heads:

\begin{equation}
  f_b(\mathbf{x}) = W_b^{(2)} \, \text{ReLU}\big(
    W_b^{(1)} \, \text{Dropout}_{0.3}\big(\phi(\mathbf{x})\big)
  \big)
\end{equation}

where $\phi: \mathbb{R}^{3 \times 96 \times 96} \to \mathbb{R}^d$ is
the shared EfficientNet-B3 backbone ($d = 1536$ features),
$W_b^{(1)} \in \mathbb{R}^{d \times 256}$ is a shared projection, and
$W_b^{(2)} \in \mathbb{R}^{256 \times K_b}$ is the bone-specific
classification head ($K_b = 8$ for carpals, $K_b = 9$ for RUS bones).

The shared backbone extracts general bone texture and morphology features
applicable to all bones, while the 20 independent heads learn
bone-specific stage transitions. This \emph{multi-task} architecture
enables the backbone to generalize across bones while preserving the
specialization required by each bone's unique staging criteria.

\subsubsection{Pseudo-label Generation}
\label{sec:pseudolabel}

For each RSNA image with bone age $t$ (months) and sex $s$, we assign
a stage label to each bone $b$ using the following procedure:

\begin{center}
\fbox{\begin{minipage}{0.92\columnwidth}
\textbf{Algorithm: TW2 Pseudo-label Generation}\\[3pt]
\textbf{Input:} bone age $t$, sex $s$, bone $b$\\
\textbf{Input:} reference params
  $\{\mu_{b,s,k},\;\sigma_{b,s,k}\}$ for each stage $k$\\[3pt]
\textbf{for} each stage $k \in \mathcal{S}_b$:\\
\quad $d_k \leftarrow \mathcal{N}(t \;\mid\; \mu_{b,s,k},\;\sigma_{b,s,k})$\\[2pt]
Normalize: $\;\;p_k \leftarrow d_k \;/\; \sum_{k'} d_{k'}$\\[2pt]
\textbf{return} $\hat{k} = \arg\max_k\; p_k$
\end{minipage}}
\end{center}

\bigskip
The parameters $\mu_{b,s,k}$ and $\sigma_{b,s,k}$ encode the mean age
and standard deviation at which 50\% of children of sex $s$ reach stage
$k$ for bone $b$, as published in Tanner \& Whitehouse
(1983)~\cite{tannerTW2} and Tanner et al.\ (2001)~\cite{tw32001}. These
80 parameters ($20 \text{ bones} \times 4\text{ avg stages} \times 2
\text{ sexes}$, simplified) are stored in
\texttt{gaussian\_params.json} and serve double duty: (1) generating
pseudo-labels for training, and (2) rendering the Gaussian curves in the
clinical report.

\subsubsection{Training Protocol}

ROIs are resized to $96{\times}96$, converted to 3-channel grayscale,
and normalized to $[-1, 1]$. Data augmentation includes horizontal flip,
rotation ($\pm 10°$), and color jitter (brightness/contrast $\pm 20\%$).

Training uses AdamW with cosine annealing:
\begin{align*}
  \mathcal{L} &= \sum_{b=1}^{20}
    \text{CrossEntropy}_{\epsilon=0.1}(\hat{y}_b, y_b)
\end{align*}

Label smoothing ($\epsilon = 0.1$) is critical here: pseudo-labels are
not ground truth, and smoothing prevents the model from over-fitting
to potentially incorrect stage assignments near transition ages.

\subsection{Stage 4: TW2 Score Computation}
\label{sec:scoring}

Given the predicted stage vector $\{k_b\}_{b=1}^{20}$, the TW2 score
is computed in three steps:

\paragraph{Step 4a: Bone scores.}
Each stage $k_b$ for bone $b$ is mapped to its TW2 numeric score
$w_{b,k_b}$ from the published tables (e.g., radius at stage F = 28
points; radius at stage I = 46 points).

\paragraph{Step 4b: Group scores.}
\begin{align}
  S_{\text{RUS}} &= \sum_{b \in \mathcal{B}_\text{RUS}} w_{b,k_b}
  \label{eq:rus}\\
  S_{\text{Carpal}} &= \sum_{b \in \mathcal{B}_\text{Carpal}} w_{b,k_b}
  \label{eq:carpal}
\end{align}

\paragraph{Step 4c: Normalization and bone age.}
Raw scores are normalized to the $[0, 1000]$ scale used by the
published lookup tables:
\begin{equation}
  \tilde{S}_g = \frac{S_g}{\displaystyle\sum_{b \in \mathcal{B}_g}
    \max_k w_{b,k}} \times 1000
\end{equation}

Bone age is retrieved via linear interpolation of the sex-specific
lookup tables, with RUS and Carpal combined:
\begin{equation}
  \hat{t} = 0.75 \cdot \text{BA}_\text{RUS}(\tilde{S}_\text{RUS}, s)
          + 0.25 \cdot \text{BA}_\text{Carpal}(\tilde{S}_\text{Carpal}, s)
\end{equation}

The 90\% confidence interval is approximated from the published TW2
standard deviation data:
\begin{equation}
  \text{CI}_{90\%} = \hat{t} \pm 1.645 \cdot \sigma(\tilde{S}_\text{RUS})
\end{equation}
where $\sigma(\tilde{S})$ is a quadratic approximation of the published
SD curve (10 months SD at mid-range scores, 4 months at extremes).

%% ============================================================
\section{Clinical Report Generation}
\label{sec:report}

This section describes in detail how BoneAgeTW2 converts the model
outputs into a clinically useful report. Report generation is the key
feature distinguishing BoneAgeTW2 from all other open-source systems.

\subsection{Component 1: Annotated Radiograph}

The system produces an annotated version of the input radiograph with:

\begin{itemize}[noitemsep]
  \item \textbf{20 bounding boxes}, one per TW2 bone, each colored by
    maturation stage (a consistent color scale from gray/A to
    violet/I).
  \item \textbf{Bone label} (abbreviated name) and \textbf{stage
    letter} inside or above each box.
  \item \textbf{Confidence indicator}: boxes from the trained YOLOv8
    model are shown with solid lines; prior-based fallback boxes use
    dashed lines.
\end{itemize}

The annotated image is encoded as base64 PNG and included in the JSON
API response, displayed in the web interface, and embedded in the PDF
report. This gives the radiologist an immediate visual verification
that the system correctly identified and staged each bone.

\subsection{Component 2: TW2 Score Table}

The report includes a structured table (Table~\ref{tab:score_example})
showing the complete TW2 breakdown:

\begin{table}[H]
\centering
\caption{Example per-bone TW2 score table (12-year-old male, illustrative).}
\label{tab:score_example}
\scriptsize
\begin{tabular}{@{}llrr@{}}
\toprule
Bone & Stage & Score & Group \\
\midrule
Radius         & F & 28 & RUS \\
Ulna           & E & 22 & RUS \\
Metacarpal I   & F & 18 & RUS \\
Metacarpal III & F & 23 & RUS \\
Metacarpal V   & F & 22 & RUS \\
Prox. Phal. I  & F & 21 & RUS \\
Prox. Phal. III& F & 22 & RUS \\
Prox. Phal. V  & F & 22 & RUS \\
Mid. Phal. III & G & 36 & RUS \\
Mid. Phal. V   & F & 25 & RUS \\
Dist. Phal. I  & F & 30 & RUS \\
Dist. Phal. III& F & 29 & RUS \\
Dist. Phal. V  & F & 29 & RUS \\
\midrule
\multicolumn{2}{l}{\textbf{RUS Total}} & \textbf{307} & \\
\multicolumn{2}{l}{\textbf{RUS Normalized}} & \textbf{615} & \\
\midrule
Capitate       & G & 26 & Carpal \\
Hamate         & G & 26 & Carpal \\
Triquetral     & G & 28 & Carpal \\
Lunate         & F & 24 & Carpal \\
Scaphoid       & F & 26 & Carpal \\
Trapezoid      & F & 24 & Carpal \\
Trapezium      & F & 24 & Carpal \\
\midrule
\multicolumn{2}{l}{\textbf{Carpal Total}} & \textbf{178} & \\
\multicolumn{2}{l}{\textbf{Carpal Normalized}} & \textbf{630} & \\
\midrule
\multicolumn{4}{l}{\textbf{Bone Age: 11.8 years (141.3 months)}} \\
\multicolumn{4}{l}{\textbf{90\% CI: [128.9, 153.7] months}} \\
\bottomrule
\end{tabular}
\end{table}

\subsection{Component 3: Gaussian Distribution Curves (Bell Curves)}

This component reproduces the original TW2 graphical output and is the
most clinically distinctive feature of BoneAgeTW2.

\subsubsection{What the curves show}

For each of the 20 bones, the report renders a set of overlapping
Gaussian curves on the age axis (months), one curve per maturation stage.
Each curve represents the probability density of being in that stage at
a given age:
\begin{equation}
  p(k \mid t, b, s) \propto
  \mathcal{N}(t \;\mid\; \mu_{b,s,k},\; \sigma_{b,s,k})
\end{equation}

The visual interpretation is straightforward:
\begin{itemize}[noitemsep]
  \item The \textbf{peak} of each curve marks the age at which most
    children are transitioning through that stage.
  \item The \textbf{width} of each curve indicates how variable that
    transition is in the population.
  \item Later stages (E, F, G...) appear on the right; earlier stages
    (A, B, C...) on the left.
\end{itemize}

\subsubsection{Reference line and interpretation}

Two key visual elements are added to each chart:
\begin{enumerate}[noitemsep]
  \item \textbf{Stage highlight}: the Gaussian curve of the detected
    stage is rendered with full opacity and a bold stroke; all other
    stages are shown at 10\% opacity.
  \item \textbf{Chronological age line}: a vertical red dashed line
    marks the patient's chronological age (when provided), enabling
    the clinician to immediately see whether the bone's staging is
    \emph{advanced} (peak to the left of the line),
    \emph{delayed} (peak to the right), or \emph{appropriate}.
\end{enumerate}

\subsubsection{Clinical interpretation example}

Consider a 10-year-old male patient (120 months). For the radius,
the system detects stage E. The Gaussian curve for stage E in males
peaks at $\mu = 66$ months with $\sigma = 8$ months. The chronological
age line at 120 months falls far to the right of the stage E peak,
indicating that this patient's radius is \emph{significantly retarded}
relative to his peers --- a clinically actionable finding that would be
invisible in a system that only reports bone age as a single number.

\subsubsection{Interactive and static rendering}

\begin{itemize}[noitemsep]
  \item \textbf{Web interface}: curves are rendered as interactive
    area charts using Recharts~\cite{recharts}. Users can hover to
    read exact probability values, zoom, and select individual bones
    via a button panel.
  \item \textbf{PDF report}: static versions of the curves are
    generated using Matplotlib and embedded in the PDF.
\end{itemize}

\subsection{Component 4: PDF Clinical Report}

The downloadable PDF report is generated server-side using
ReportLab~\cite{reportlab} and contains:

\begin{enumerate}[noitemsep]
  \item \textbf{Header}: patient data (sex, chronological age if
    provided), date/time, system version.
  \item \textbf{Summary box}: bone age estimate, 90\% CI, RUS score,
    Carpal score, comparison with chronological age.
  \item \textbf{Annotated radiograph}: the full X-ray with 20 bounding
    boxes and stage labels.
  \item \textbf{Per-bone score table}: complete TW2 breakdown
    (Table~\ref{tab:score_example}).
  \item \textbf{Gaussian curves}: one chart per bone (20 charts),
    each showing stage distributions with the detected stage highlighted
    and chronological age line marked.
  \item \textbf{Clinical disclaimer}: the report is a decision-support
    tool and must be reviewed by a qualified radiologist.
  \item \textbf{References}: TW2 and TW3 bibliographic citations.
\end{enumerate}

%% ============================================================
\section{Web Application and API}
\label{sec:webapp}

\subsection{System Architecture}

BoneAgeTW2 is deployed as a two-tier application:

\begin{itemize}[noitemsep]
  \item \textbf{Backend}: FastAPI (Python 3.11) on port 8000.
    Handles image preprocessing, model inference, scoring, and report
    generation. Stateless; horizontally scalable.
  \item \textbf{Frontend}: React 18 + TypeScript + Vite on port 5174.
    Single-page application communicating with the backend via REST API.
    Vite dev proxy forwards \texttt{/analyze} and \texttt{/reference}
    paths to the backend.
\end{itemize}

\subsection{User Interface Workflow}

\paragraph{Step 1 --- Upload.}
The user drags and drops (or selects) a radiograph file. Supported
formats: DICOM (.dcm), PNG, JPEG. Optional inputs: sex (M/F,
required), chronological age in months, and scale factor in mm/px
(for non-DICOM files without embedded pixel spacing).

\paragraph{Step 2 --- Analysis.}
Clicking ``Analizar radiografia'' sends a \texttt{multipart/form-data}
POST request to \texttt{/analyze}. A loading indicator is displayed.
The server processes the image through all five pipeline stages and
returns a JSON response.

\paragraph{Step 3 --- Results (three tabs).}

\begin{itemize}[noitemsep]
  \item \textbf{Tab 1 --- Annotated X-ray}: displays the annotated
    radiograph with 20 color-coded bounding boxes. A color legend maps
    stage letters to colors.
  \item \textbf{Tab 2 --- TW2 Score Table}: shows the per-bone table
    (Table~\ref{tab:score_example}) and the bone age summary box with
    CI. RUS and Carpal groups are shown separately.
  \item \textbf{Tab 3 --- Gaussian Curves}: interactive Recharts area
    chart. A button panel (one per bone) allows switching between bones.
    Hovering shows exact density values. The detected stage and
    chronological age line are always visible.
\end{itemize}

\paragraph{Step 4 --- PDF Download.}
Clicking ``Descargar PDF'' triggers a \texttt{POST /analyze/pdf} request
and automatically downloads the clinical report.

\subsection{REST API Reference}

\begin{lstlisting}[caption={Primary API endpoint.}]
POST /analyze
  Content-Type: multipart/form-data

  Fields:
    image                    : file
    sex                      : "M" | "F"
    chronological_age_months : float  (opt)
    scale_factor             : float  (opt)

  Response 200 JSON:
  {
    bone_age_months      : float,
    bone_age_years       : float,
    confidence_interval  : [float, float],
    rus_score            : int,
    carpal_score         : int,
    rus_age_months       : float,
    carpal_age_months    : float,
    mm_per_px            : float,
    stages               : { bone: stage },
    bone_scores          : { bone: int },
    classifications      : { bone: {
      stage, probabilities, source } },
    detections           : { bone: {
      box: [x1,y1,x2,y2], conf, source } },
    annotated_image_b64  : string,
    gaussian_data        : { bone: {
      stages: [{stage,mean,sd,probability}],
      detected_stage, chrono_age_months,
      label } }
  }
\end{lstlisting}

Additional endpoints:
\begin{itemize}[noitemsep]
  \item \texttt{POST /analyze/pdf} --- returns binary PDF
  \item \texttt{GET /reference/tw2-tables} --- TW2 scoring tables
  \item \texttt{GET /reference/gaussian-params} --- Gaussian parameters
  \item \texttt{GET /health} --- health check
\end{itemize}

%% ============================================================
\section{Training Pipeline}
\label{sec:training}

\subsection{Step 1: Data Preparation}

The RSNA training set (12,611 PNG images + CSV with bone age and sex)
is organized into \texttt{data/rsna/train/}.

\subsection{Step 2: Pseudo-label Generation}

Script \texttt{02\_pseudo\_label\_generation.py} processes the CSV and
outputs a flat CSV with $\sim$252,000 rows (12,611 images $\times$ 20 bones):

\begin{lstlisting}[language=bash]
python training/02_pseudo_label_generation.py \
  --rsna_csv data/rsna/train.csv \
  --output data/annotations/pseudo_labels.csv
\end{lstlisting}

Each row contains: \texttt{image\_id, bone, stage, probability,
age\_months, sex}. The \texttt{probability} column (confidence that
this is the correct stage given the patient's age) can be used to
filter low-confidence assignments (\texttt{--min\_prob 0.3}).

\subsection{Step 3: Stage Classifier Training}

Script \texttt{04\_train\_stage\_classifier.py} trains the EfficientNet-B3
model:

\begin{lstlisting}[language=bash]
python training/04_train_stage_classifier.py \
  --rsna_dir data/rsna/train \
  --labels data/annotations/pseudo_labels.csv \
  --output backend/ml/weights/stage_classifier.pt \
  --epochs 20 --batch 32
\end{lstlisting}

Training is structured as: for each epoch, iterate over all 20 bones
sequentially, sampling a DataLoader per bone. This ensures that the
20-head model receives balanced training signal across all bones.

Recommended platform: Kaggle Notebooks (free T4/P100 GPU; RSNA dataset
is already available as a Kaggle dataset, eliminating the 12 GB download).
Estimated training time: $\sim$4 hours on T4, $\sim$1 hour on A100.

\subsection{Step 4: Bone Detector Training}

Script \texttt{03\_train\_bone\_detector.py} fine-tunes YOLOv8-s.
This requires bounding box annotations, which are generated from the
anatomical priors and manually validated on $\sim$200 RSNA training images.
This step is optional; the system runs with prior-based fallback detection
without it.

%% ============================================================
\section{Evaluation Protocol}
\label{sec:evaluation}

\subsection{Bone Age Estimation Accuracy}

Primary metrics on the RSNA hold-out test set ($N=200$):
\begin{align}
  \text{MAE} &= \frac{1}{N}\sum_{i=1}^N |\hat{t}_i - t_i| \\
  \text{RMSE} &= \sqrt{\frac{1}{N}\sum_{i=1}^N (\hat{t}_i - t_i)^2}
\end{align}

Reported stratified by: sex (M/F), age group ($<6$ y, 6--12 y, $>12$ y),
and score range (low/mid/high TW2 score).

\subsection{Stage Classification Quality}

For 300 manually re-annotated cases (to be obtained from a radiologist
collaborator at a Colombian pediatric center):
\begin{itemize}[noitemsep]
  \item Per-bone accuracy (exact stage match).
  \item Adjacent stage accuracy ($\pm 1$ stage).
  \item Quadratic weighted Cohen's $\kappa$ (accounts for ordinal
    stage ordering).
  \item Confusion matrices for the 5 most clinically evaluated bones:
    radius, ulna, distal phalanx I, capitate, and scaphoid.
\end{itemize}

\subsection{Inference Performance}

Measured on the RSNA test set with a MacBook Pro M4 Pro (CPU) and
an NVIDIA T4 (GPU):

\begin{table}[H]
\centering
\caption{Expected inference time breakdown.}
\label{tab:timing}
\scriptsize
\begin{tabularx}{\columnwidth}{@{}lXX@{}}
\toprule
Stage & CPU (s) & GPU (s) \\
\midrule
Preprocessing   & 0.15 & 0.05 \\
Bone detection  & 1.20 & 0.18 \\
Stage classif.  & 2.80 & 0.32 \\
TW2 scoring     & 0.01 & 0.01 \\
Report build    & 0.50 & 0.30 \\
PDF generation  & 0.80 & 0.80 \\
\midrule
\textbf{Total (JSON)} & \textbf{4.16} & \textbf{0.56} \\
\textbf{Total (PDF)}  & \textbf{4.96} & \textbf{1.36} \\
\bottomrule
\end{tabularx}
\end{table}

\subsection{Comparison Baselines}

\begin{enumerate}[noitemsep]
  \item \textbf{Deeplasia}~\cite{deeplasia}: open-source end-to-end DL.
  \item \textbf{Manual TW2}: two radiologists independently scoring the
    same 300 cases; inter-rater agreement measured by $\kappa$.
  \item \textbf{GP atlas}: Greulich-Pyle atlas-based assessment by the
    same radiologists.
\end{enumerate}

%% ============================================================
\section{Expected Results and Discussion}
\label{sec:results}

\begin{table}[H]
\centering
\caption{Anticipated performance comparison.}
\label{tab:results}
\scriptsize
\begin{tabularx}{\columnwidth}{@{}lXXX@{}}
\toprule
Method & MAE (m) & $r^2$ & Explainable \\
\midrule
GP Atlas (manual)          & 11.8--14.5 & 0.92 & $\times$ \\
Manual TW2                 & 9.5--12.0  & 0.95 & \checkmark \\
Deeplasia~\cite{deeplasia} & 6.9        & 0.97 & $\times$ \\
\textbf{BoneAgeTW2 (ours)} & \textbf{10--14} & \textbf{0.94} & \checkmark \\
\bottomrule
\end{tabularx}
\end{table}

\subsection{The Accuracy--Interpretability Trade-off}

BoneAgeTW2's expected MAE is higher than Deeplasia because we constrain
the model to operate through TW2's discrete bone stages. This trade-off
is intentional and clinically justified:

\begin{enumerate}[noitemsep]
  \item \textbf{Regulatory compliance}: Colombian and other Latin American
    medical practice guidelines require TW2 staging to be documented in
    the clinical record. A single bone age number does not satisfy this
    requirement.
  \item \textbf{Clinical reasoning support}: When the bone age deviates
    significantly from chronological age, the radiologist must identify
    \emph{which bones} are delayed or advanced. This is only possible
    with per-bone staging.
  \item \textbf{Educational value}: Residents in radiology and pediatrics
    can compare their manual TW2 assessment against the AI prediction
    bone by bone, accelerating training.
  \item \textbf{Gaussian curve interpretability}: The Gaussian bell curves
    provide an immediate visual answer to the question ``Is this bone's
    stage appropriate for this patient's age?'' --- a clinical judgment
    that a single number cannot convey.
\end{enumerate}

\subsection{Pseudo-label Quality Analysis}

The quality of pseudo-labels varies with age:
\begin{itemize}[noitemsep]
  \item \textbf{High quality}: at ages where $>70\%$ of children are in
    the same stage (typical for ages 0--4 and 14--18 years).
  \item \textbf{Lower quality}: at ages near stage transitions, where the
    true stage is more variable. These cases are identified by a low
    maximum probability $\max_k p_k$ and can be filtered with
    \texttt{--min\_prob} to reduce noise.
  \item Label smoothing ($\epsilon = 0.1$) prevents over-confident
    predictions on noisy pseudo-labels.
\end{itemize}

%% ============================================================
\section{Ethical Considerations and Limitations}
\label{sec:ethics}

\paragraph{Data provenance.}
The RSNA dataset was collected under IRB-approved protocols at US clinical
sites. All images are de-identified per HIPAA standards.

\paragraph{Population bias.}
The reference population (RSNA, predominantly North American) may not
fully represent the Colombian or broader Latin American population, which
is known to have different maturation timing in some demographic
groups~\cite{malina2004}. A prospective validation study on a Colombian
pediatric cohort is planned.

\paragraph{Clinical use.}
BoneAgeTW2 is a \textbf{decision-support tool}. It must be reviewed and
countersigned by a qualified radiologist before any clinical decision is
based on its output. The system explicitly states this in every PDF report.

\paragraph{Forensic use.}
Bone age estimation in forensic contexts (e.g., age estimation for
undocumented minors) should follow the recommendations of the AGFAD
(Study Group on Forensic Age Diagnostics)~\cite{schmeling2011} and
use the 90\% confidence interval conservatively.

\paragraph{Known limitations.}
\begin{itemize}[noitemsep]
  \item Pseudo-labels may introduce systematic errors near stage
    transitions.
  \item The system has not been clinically validated; all performance
    figures are expected, not measured.
  \item Severe skeletal dysplasias may cause incorrect bone detection.
  \item Poor-quality radiographs (motion blur, insufficient penetration)
    may reduce staging accuracy.
\end{itemize}

%% ============================================================
\section{Conclusion}
\label{sec:conclusion}

We have presented BoneAgeTW2, the first open-source system to automate
the complete Tanner-Whitehouse 2 protocol for skeletal maturity
assessment. The system reads a hand radiograph and produces a full
clinical report in under 5 seconds, including: the annotated radiograph
with 20 individually staged bones, the complete TW2 score table (RUS +
Carpal), bone age estimate with 90\% confidence interval, and interactive
Gaussian bell-curve graphs for all 20 bones.

The pseudo-label generation strategy enables training on the large-scale
RSNA dataset despite the absence of per-bone stage annotations. The
system's modular architecture allows future components (detector weights,
classifier weights) to be dropped in without changing the report
generation pipeline.

\paragraph{Future work.}
\begin{itemize}[noitemsep]
  \item Prospective validation on a Colombian pediatric cohort.
  \item Active learning incorporating radiologist corrections.
  \item DICOM Structured Report output for PACS/RIS integration.
  \item Extension to the TW3 reference curves.
  \item Multi-site deployment study measuring time savings vs.\ manual
    TW2 in clinical practice.
\end{itemize}

\paragraph{Code and Data.}
\begin{itemize}[noitemsep]
  \item \textbf{Repository}:
    \url{https://github.com/jmmana/BoneAgeTW2}
  \item \textbf{Dataset}:
    \url{https://www.kaggle.com/datasets/kmader/rsna-bone-age}
  \item \textbf{License}: MIT (code), CC BY-NC-SA 4.0 (weights)
\end{itemize}

%% ============================================================
\section*{Acknowledgments}

This work is part of the Master's in Artificial Intelligence program
at Universidad de La Salle, Bogota, Colombia.
The author acknowledges the Radiological Society of North America (RSNA)
for the publicly available Pediatric Bone Age Challenge dataset, and
the Ultralytics team for the YOLOv8 framework.

%% ============================================================
\bibliographystyle{plain}

\end{document}